\documentclass[10pt,twocolumn,letterpaper]{article}

\usepackage[pagenumbers]{cvpr} 

\definecolor{cvprblue}{rgb}{0.21,0.49,0.74}
\usepackage[pagebackref,breaklinks,colorlinks,allcolors=cvprblue]{hyperref}

\usepackage{bm}
\usepackage{multirow}\newtheorem{corollary}{Corollary}
\newtheorem{Thm}{Theorem}

\usepackage{amsmath}

\def\confName{CVPR}
\def\confYear{2025}

\title{Rotation-Equivariant Self-Supervised Method in Image Denoising}

\author{Hanze Liu$^{1}$ \hspace{2em}  Jiahong Fu$^{1}$ \hspace{2em}  Qi Xie$^{1,\dagger}$ \hspace{2em}  Deyu Meng$^{1,2,3}$\\
$^{1}$  Xi’an Jiaotong University, Xi'an, China\\
$^{2}$ Pengcheng Laboratory, Shenzhen, China\\
$^{3}$ Macau University of Science and Technology, Taipa, Macao \\
}

\begin{document}
\maketitle
\def\thefootnote{}\footnotetext[1]{$^\dag$ Corresponding author.}
\def\thefootnote{\arabic{footnote}}
\begin{abstract}
Self-supervised image denoising methods have garnered significant research attention in recent years, for this kind of method reduces the requirement of large training datasets.
Compared to supervised methods, self-supervised methods rely more on the prior embedded in deep networks themselves. As a result, most of the self-supervised methods are designed with Convolution Neural Networks (CNNs) architectures, which well capture one of the most important image prior, translation equivariant prior. Inspired by the great success achieved by the introduction of translational equivariance, in this paper, we explore the way to further incorporate another important image prior. 
Specifically, we first apply high-accuracy rotation equivariant convolution to self-supervised image denoising. Through rigorous theoretical analysis, we have proved that simply replacing all the convolution layers with rotation equivariant convolution layers would modify the network into its rotation equivariant version.
To the best of our knowledge, this is the first time that rotation equivariant image prior is introduced to self-supervised image denoising at the network architecture level with a comprehensive theoretical analysis of equivariance errors, which
offers a new perspective to the field of self-supervised image denoising.
Moreover, to further improve the performance, we design a new mask mechanism to fusion the output of rotation equivariant network and vanilla CNN-based network, and construct an adaptive rotation equivariant framework. 
Through extensive experiments on three typical methods, we have demonstrated the effectiveness of the proposed method. The code is available at: \href{https://github.com/liuhanze623/AdaReNet}{https://github.com/liuhanze623/AdaReNet}.
\end{abstract}     
\vspace{-4mm}
\section{Introduction}
\label{sec:intro}

During the process of capturing and transmitting images, unexpected noises are frequently introduced~\cite{henkelman1985measurement,liu2022shaking}. Such noise can severely degrade image quality and disrupt subsequent image processing tasks. Image denoising is a technique designed to address the standard inverse problem in image processing and is widely utilized across various fields. 
With the rapid advancement of deep learning (DL), denoisers based on different deep networks have achieved outstanding results. 
\setlength{\textfloatsep}{7pt} 
\begin{figure}[t]
  \centering
   \includegraphics[width=\linewidth]{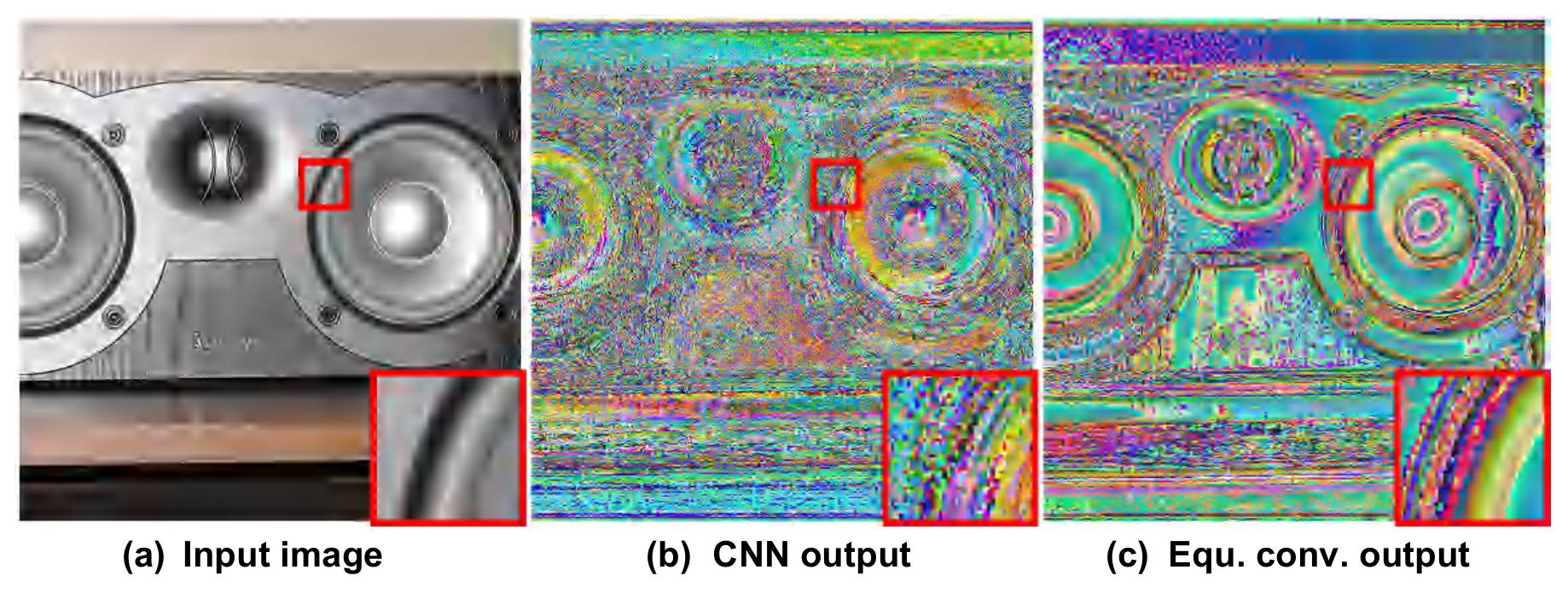}
   \caption{Illustration of the output feature map of a typical image obtained by standard CNN and our used rotation equivariant convolution neural network. Both networks are initialized randomly.}
   \label{fig:abc}
\end{figure}

Early, it was believed that the success of deep learning is primarily due to the exhaustive utilization of training data. Thus most early DL-based image denoising are in supervised manner~\cite{zhang2017beyond,zhang2018ffdnet,anwar2019real,chang2020spatial,vemulapalli2016deep, kim2020transfer,park2019densely,guo2019toward}, where models are trained on extensive datasets that include pairs of clean and noisy images, thereby learning the transformation from noisy to clean images. However, in reality, compiling such comprehensive datasets is both costly and time-intensive, posing substantial challenges~\cite{anaya2018renoir,plotz2017benchmarking,xu2018real}. 

Self-supervised approaches thus achieved significant research attention in recent years. These approaches rely less on extensive supervised datasets, but pay more attention to the full utilization of the prior information inherent in deep networks \cite{lehtinen2018noise2noise,krull2019noise2void,batson2019noise2self,moran2020noisier2noise,xu2020noisy,cha2019gan2gan,huang2021neighbor2neighbor,pang2021recorrupted,wang2022blind2unblind,zhang2023unleashing}.
Notably, Lehtinen \etal introduced the innovative self-supervised learning approach called Noise2Noise~\cite{lehtinen2018noise2noise}, which enables training on pairs of noisy images depicting the identical scene. ~\cite{lehtinen2018noise2noise} was the first self-supervised algorithm that achieved performance comparable to supervised image denoising. Other methods involve training denoising models on single noisy images by designing blind-spot networks~\cite{krull2019noise2void,batson2019noise2self,wu2020unpaired,krull2020probabilistic}, while some approaches devise strategies to generate training image pairs from noisy images~\cite{moran2020noisier2noise,xu2020noisy,cha2019gan2gan,huang2021neighbor2neighbor,pang2021recorrupted} and have achieved excellent results.

Nowadays, it has become common sense that the prior information inherent in deep networks plays an important role in improving the performance of DL-based methods in image processing tasks \cite{weiler2018learning,shen2020pdo,xie2022fourier,fu2024rotation}. The most classic example should be the CNN, which well captures the translation-equivariant prior of natural images. In CNNs, shifting an input image of CNN is equivalent to shifting all of its intermediate feature maps and the output image. Compared with fully-connected neural
networks, this translational equivariance property brings in rational weight sharing, makes the network parameters being used more efficiently, and thus leads to substantially better performance. 
Inspired by the significant success brought about by the introduction of translational equivariance, another essential image prior (As illustrated in \cref{fig:abc}), rotation-equivariant prior has to be taken into consideration very recently. Rotation-equivariant CNNs have been applied and achieved notable performance improvement in multiple supervised image processing tasks \cite{xie2022fourier,fu2024rotation}.

Compared with supervised learning, self-supervised learning approaches indeed depend more on the prior knowledge embedded within the network, for the information that can be learned from the dataset is less.  
As a result, most current self-supervised image denoising methods are all based on the CNN architectures, which is largely due to the reliance on the prior of translational equivariance. 
Therefore, it holds even greater significance to 
incorporate rotational equivariant design into the deep networks for self-supervised methods. 

However, there are two critical issues that need to be addressed when introducing equivariant priors into self-supervised image denoising.
On the one hand, although, Xie \etal have recently constructed a rotation equivariant CNN suitable for image processing \cite{xie2022fourier} and Fu \etal further analyzed the global rotation equivariance error of the entire network \cite{fu2024rotation}, current rotation equivariant designs for image processing are usually in ResNet\cite{he2016deep} structure without upsampling or downsampling. For example, for the super-resolution network in \cite{xie2022fourier} where upsampling module is inevitable, only the portion of the network before upsampling is rotation equivariant and the equivariance for the entire network can not be guaranteed.
Since self-supervised networks often utilize U-Net structures, which include multiple upsampling and downsampling modules, it is imperative to first explore the impact of these modules on rotation equivariance of the network.

On the other hand, rotation equivariant design for deep networks inevitably introduces parameter sharing and convolution kernel parameterization, which usually degrades the representation accuracy of the network. However, image denoising tasks often have high requirements for the network’s representation accuracy, since important high-frequency components need to be reconstructed. Besides, not all areas of natural images strictly comply with rigid rotation equivariance, indiscriminately adopting a rotation equivariant network for the entire image can often be detrimental to the reconstruction performance of images.

This study primarily focuses on integrating rotation equivariant prior to existing self-supervised denoising techniques while solving the aforementioned issues. The key contributions can be summarized as follows:
\begin{itemize}
\item We explore the way for introducing rotation equivariant prior into self-supervised image denoising frameworks at the network architecture level. Particularly, for the first time, we rigorously analyze the impact of upsampling and downsampling on equivariant networks theoretically. The equivariant errors of upsampling and downsampling layer indeed approach zero when the resolution of the input image increases, though the approach rate ($O(h)$, $h$ denoted the mesh size of image) is slower than equivariant convolutional layer (whose equivariant error approaches zero at a rate of $O(h^2)$). 
Then, by taking U-Net as an example, we further analyze the rotation equivariant error of the entire network, showing that by simply replacing all the convolution layers with rotation equivariant convolutions \cite{xie2022fourier}, we can indeed achieve a reliable rotation equivariant network for self-supervised image denoising. 
      
\item We have further developed an adaptive rotation equivariant network to enhance the representation accuracy. Specifically, we design a fusion module for integrating the advantages of both rotation-equivariant and non-equivariant networks. The module can automatically determine which regions of the image would benefit more from a rotation-equivariant network compared to a normal CNN-based network. This design provides greater flexibility and yields improved denoising results.
\item We conducted comprehensive experiments across various self-supervised denoising methods, demonstrating the effectiveness of integrating rotation-equivariant image priors into neural networks for self-supervised techniques. Our approach provides a novel perspective in the field of self-supervised image denoising.
\end{itemize}

\section{Related Work and Prior Knowledge}
\label{sec: Related Work}

\subsection{Image Denoising}

\noindent\textbf{Non-learning Denoising Methods:} 
Conventional denoising methods predominantly depend on the statistical characteristics of images and mathematical modeling, often employing image priors instead of learned denoisers. Notably, NLM~\cite{buades2005non,buades2011non} and BM3D~\cite{dabov2007image,makinen2019exact} have been proposed that can effectively remove noise based on the exploitation of image self-similarity. WNNM~\cite{gu2014weighted,gu2017weighted} treated image denoising as a low-rank matrix approximation problem and achieved great performance.

\noindent\textbf{Supervised Denoising Methods:} 
The majority of DL-based denoising algorithms are supervised~\cite{zhang2018ffdnet,zhang2017beyond,guo2019toward,yue2020dual,anwar2019real,chang2020spatial}. DnCNN~\cite{zhang2017beyond} employed an end-to-end training strategy to learn the mapping between noisy images and the residual components, markedly improving denoising efficacy. FFDNet~\cite{zhang2018ffdnet} proposed a versatile denoising architecture by incorporating the noise level as a network parameter.
Nevertheless, to train a proficient model, an extensive collection of paired noisy and clean images is necessary to comprehensively capture the range of image contents and noise variations~\cite{quan2021image,zamir2020learning,anwar2019real,jia2019focnet,guo2019toward,lefkimmiatis2018universal}. The acquisition of such training datasets can be prohibitively costly and challenging, and in the case of certain medical images, it may be virtually unattainable.

\noindent\textbf{Self-supervised Denoising Methods:} Self-supervised deep learning methods have garnered significant interest due to their independence from clean reference images~\cite{lehtinen2018noise2noise,krull2019noise2void,batson2019noise2self,moran2020noisier2noise,xu2020noisy,cha2019gan2gan,huang2021neighbor2neighbor,pang2021recorrupted,wang2022blind2unblind,zhang2023unleashing}. Noise2Noise~\cite{lehtinen2018noise2noise}, proposed by Lehtinen \etal, was the first self-supervised method that achieved performance comparable to supervised methods only using paired noisy images. 
Noise2Void~\cite{krull2019noise2void} and Noise2Self~\cite{batson2019noise2self} used a single noisy image and employed a blind-spot network to predict the clean pixel values based on neighboring pixels, thereby avoiding collapse to an identity mapping. ~\cite{moran2020noisier2noise,xu2020noisy,cha2019gan2gan,huang2021neighbor2neighbor,pang2021recorrupted} devise strategies to generate training image pairs from noisy images to train a network. 
R2R~\cite{pang2021recorrupted} has demonstrated that the cost function defined on the noisy/noisy image pairs generated by this method is statistically equivalent to the supervised method. Single-image denoising methods~\cite{ulyanov2018deep,quan2020self2self} capitalize on the statistical properties of the image itself, showing powerful denoising capabilities without extensive training datasets. These methods lack clean images for supervision and rely more on the prior knowledge embedded within the network, which is the reason for the success of these CNN-based self-supervised image denoising methods. Inspired by this, we explore the way to further incorporate rotation equivariant image prior to these methods with rigorous theoretical analysis.

\subsection{Rotation Equivariance}
Equivariant Convolution Neural Networks (ECNNs) have drawn substantial interest within the field of computer vision these years~\cite{cohen2016group,hoogeboom2018hexaconv,zhou2017oriented,marcos2017rotation,worrall2017harmonic,weiler2018learning,weiler2019general,shen2020pdo,shen2021pdo}. 
Their key strength stems from their ability to effectively handle image rotations through architectural design, which significantly enhances the model's generalization and robustness.

Cohen \etal introduced Group Equivariant Convolutional Networks (G-CNNs)~\cite{cohen2016group} and first integrated $\frac{\pi}{2}$ degree rotation equivariance into the neural network. Recently, the filter parametrization technique has been widely employed 
and Weiler \etal used harmonics as steerable filters to achieve exact equivariance~\cite{weiler2018learning,weiler2019general}. However, a notable limitation of these filter parameterization techniques is the inaccuracy in their representation, which significantly affects low-level vision tasks. 
Fortunately, Xie \etal addressed this issue by proposing Fourier series expansion-based filter parametrization, which has relatively high expression accuracy~\cite{xie2022fourier}. 
The proposed Fconv exhibits precise equivariance in the continuous domain, degrading to approximation only after discretization. 

Equivariant convolutions, distinct from data augmentation techniques, inherently integrate rotational symmetry image priors into the network architecture, thereby guaranteeing the network's inherent equivariance and offering superior interpretability and generalizability. 
However, in the U-Net structure, there is no theoretical guarantee for the design of rotation equivariance.
To the best of our knowledge, this study represents the first application of rotation equivariant convolution in the field of self-supervised image denoising and offers theoretical assurance of equivariance for networks based on the U-Net architecture.

\subsection{Prior Knowledge about Equivariance}
Equivariance to a transformation indicates that the application of a transformation to the input results in a corresponding, predictable transformation of the output~\cite{shen2020pdo,weiler2019general,xie2022fourier}. Concretely, consider a mapping $\Psi$ that transforms the input feature space to the output feature space, and let $G$ denote a set of transformations. For any $g \in G$, the following relationship holds:
\vspace{-2mm}
\begin{equation}
\Psi\left[\pi_{g}[f]\right]=\pi'_{g}\left[\Psi[f]\right],
\end{equation}
where $f$ represents any feature map within the input feature space, and $\pi_{g}$ and $\pi'_{g}$ describe the action of the transformation $g$ on the input and output features, respectively.
\section{Proposed Method}
\label{Proposed Method}

\subsection{ECNNs for Self-supervised Image Denoising}

Self-supervised image denoising methods lack clean images for supervision and rely more on the prior knowledge embedded within the network. Therefore, introducing rotation equivariant image prior to this field is reasonable, which can be achieved by ECNNs~\cite{xie2022fourier}. The U-Net can effectively restore high-frequency details in images with an encoder-decoder architecture, making it widely used in this task. Consequently, to construct an equivariant self-supervised denoising network, we need to discuss the impact of the essential upsampling and downsampling operators in the U-Net network on equivariance.

In the subsequent section, we first present some notations and concepts, and then theoretically analyze the impact of upsampling and downsampling operators on equivariant networks. Furthermore, we deduce the equivariant error for the complete U-Net network and provide theoretical guarantees for the implementation of equivariance in U-Net architectures.

\begin{figure}[t]
  \centering
  \includegraphics[width=\linewidth]{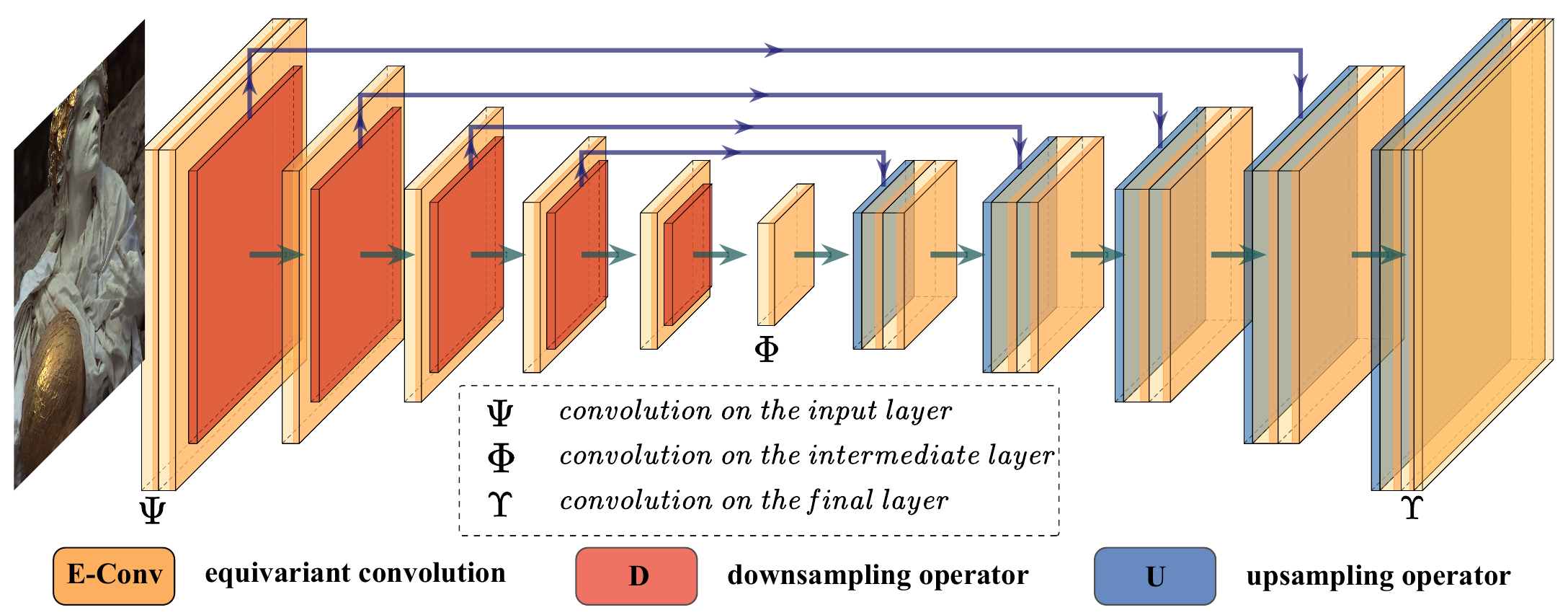}
  \caption{The network architecture of the equivariant N2N method. The network can be divided into multiple upsampling and downsampling blocks. Each downsampling block (DB) consists of one E-Conv layer and a downsampling operator, while each upsampling block (UB) is composed of an upsampling operator and two E-Conv layers.}
  \label{fig:N2N_UNet}
\end{figure}

\subsubsection{Notations and Concepts}
We first introduce some necessary notations and preliminaries as follows. 

We consider the equivariance on the orthogonal group  $O(2)$. Formally,
$O(2)=\{A\in\mathbb{R}^{2\times 2}|A^TA = I_{2\times  2}\}$, which contains all rotation and reflection matrices. Without ambiguity, we use $A$ to parameterize $O(2)$. The Euclidean group $E(2) = \mathbb{R}^2 \rtimes O(2)$ ($\rtimes$ is a semidirect-product), whose element is represented as $(x, A)$. Restricting the domain of $A$ and $x$, we can also use this representation to parameterize any subgroup of $E(2)$. In practice, the subgroup is usually assumed to contain $t$ rotations with $\frac{2\pi}{t}$ degree for an integer $t\in \mathbb{N}_+$.

An image $I \in R^{n\times n}$ is viewed as a two-dimensional
 discretization of a smooth function $r: \mathbb{R}^2\to\mathbb{R}$, at the cell-center of a regular grid with $n\times n$ cells, \ie., for $i, j = 1,2, \cdots, n$,
\vspace{-1mm}\begin{equation}\label{I}
  I_{ij} = r(x_{ij}),
\end{equation}
where $x_{ij} \!=\! \left(\left(i\!-\!\frac{n\!+\!1}{2}\right)h, \left(j\!-\!\frac{n\!+\!1}{2}\right)h\right)^T\!$ ,  $h$ is the mesh size.

An intermediate feature map $F\in \mathbb{R}^{n\times n \times t}$ in equivariant networks is a multi-channel  tensor, which can be viewed as the discretization of a continuous function defined on $\tilde{E} = \mathbb{R}^2\rtimes S$, where $S$ is a subgroup
 of $O(2)$ and $t$ is the number of elements in $S$. Formally,   $F$ can be represented as a three-dimensional grid tensor sampled from a smooth function $e: \mathbb{R}^2\times S\to \mathbb{R}$, \ie., for $i, j = 1,2, \cdots, n$,
\vspace{-1mm}\begin{equation}\label{F}
  F_{ij}^A = e(x_{ij}, A),
\end{equation}\vspace{-1mm}
where $x_{ij}$ is defined in \eqref{I} and $A\in S$.

With above notations, the transformations on the input and feature maps can be mathematically formulated. Specifically, in the continuous domain, for an input $r\in C^\infty(\mathbb{R}^2)$ and feature map $e\in C^\infty(E(2))$, the transformation $\tilde{A}\in O(2)$ acts on $r$ and $e$ respectively by:
\begin{equation}\label{piR}
\begin{split}
  \pi^R_{\tilde{A}}[r](x) &= r(\tilde{A}^{-1}x), \forall x\in\mathbb{R}^2,\\
  \pi^E_{\tilde{A}}[e](x,A) &= e(\tilde{A}^{-1}x,\tilde{A}^{-1}A), \forall (x,A)\in E(2).
\end{split}
\end{equation}
In particular, if $A_{\theta}\in O(2)$ is the rotation matrix $\begin{bmatrix}\cos\theta,\sin\theta\\-\sin\theta,\cos\theta\end{bmatrix}$, then the corresponding rotation operators can be expressed by $\pi^R_{\theta}$ and $\pi^E_{\theta}$.

Besides, in the discrete domain, we can also define the transformation $\tilde{A}\in S$ on the input image and feature map as followings:
\begin{equation}\label{Pi_d}
\begin{split}
     \left(\tilde{\pi}_{\tilde{A}}^{R}(I)\right)_{ij} &= \pi_{\tilde{A}}^{R}[r](x_{ij}), \\
     \left(\tilde{\pi}_{\tilde{A}}^{E}(F)\right)_{ij}^A &= \pi_{\tilde{A}}^{E}[e](x_{ij},A), \\
     \forall i,j = 1,&2,\cdots,n,  A \in S.
\end{split}
\end{equation}
Similarly, rotation operators can be denoted as $\tilde{\pi}^R_{\theta}$ and $\tilde{\pi}^E_{\theta}$.

\subsubsection{Equivariance of Downsampling and Upsampling}

As shown in \cref{fig:N2N_UNet}, the U-Net architecture commonly used in self-supervised image denoising typically integrates upsampling and downsampling operators. In specific applications, these layers will inevitably affect the equivariance of the network. 
Recent studies \cite{fu2024rotation, xie2022fourier} have analyzed the equivariance properties of the group convolution layer. However, the effects of upsampling and downsampling on network equivariance remain unexplored.

First of all, We provide the definitions of commonly used downsampling methods in the continuous domain.

\noindent\textbf{Maxpooling Downsampling.} Maxpooling is a commonly used downsampling method in CNNs, which reduces the spatial dimensions of feature maps by sliding a fixed-size window over the feature map and selecting the maximum value within each region as the output~\cite{lecun1989backpropagation}. In the continuous domain, we can define maxpooling operator $M(\cdot)$ as follows,
\begin{equation}
[M(F)](x,A)=max_{\Omega_{x}}F_{ij}^{A},
\end{equation}
where $x=[x_1,x_2]^T\in\mathbb{R}^2$ denotes the spatial coordinates, and $x_{1} \in [i,i+1]$, $x_{2} \in [j,j+1]$, $\Omega_{x}=\{(i,j),(i+1,j),(i,j+1),(i+1,j+1)\}$.

\noindent\textbf{Stride Downsampling.} Stride Downsampling is also a widely used downsampling operator which reduce the size of the feature map by adjusting the stride of the convolution operation~\cite{lecun1998gradient}. In the continuous domain, we can define stride downsampling operator $S(\cdot)$ as follows,
\begin{equation}
[S(F)](x,A)=F_{i,j+1}^{A},
\end{equation}
where $x=[x_1,x_2]^T\in\mathbb{R}^2$ denotes the spatial coordinates, and $x_{1} \in [i,i+1]$, $x_{2} \in [j,j+1]$. $\Omega_{x}=\{(i,j),(i+1,j),(i,j+1),(i+1,j+1)\}$.

For the above two common downsampling operators, it is of great significance to analyze their impact on the equivariant network. Therefore, we constructed their equivariant errors under the rotational equivariant structure.

\begin{Thm}
Assume that a feature map $F \in \mathbb{R}^{n \times n \times t}$ is discretized from the smooth function $e: \mathbb{R}^2 \times S \rightarrow \mathbb{R}$, $|S|=t$, the mesh size is $h$, $D(\cdot)$ is the downsampling operator. 
If for any $A,B \in S, x \in \mathbb{R}^2$, the following conditions are satisfied:
\begin{equation}
\parallel\nabla e(x,A)\parallel\leq G,
\end{equation}
then the following results are satisfied:
\begin{equation}
|D\left[\tilde{\pi}_{B}^E\right]\left(F\right)\left(x,A\right) - \pi_{B}^E\left[D\left(F\right)\right]\left(x,A\right)| \leq 2\sqrt{2}Gh. 
\end{equation}
\end{Thm}

The downsampling operator $D(\cdot)$ can be either $M(\cdot)$ or $S(\cdot)$. Theorem 1 reveals that the equivariant error of the downsample operator is primarily influenced by the mesh size $h$, and it indeed approach zero when the resolution of the input image increases with the approach rate $O(h)$.

Then, we provide the definitions of commonly used upsampling methods in the continuous domain.

\noindent\textbf{Nearest Neighbor Upsampling.} Nearest neighbor interpolation is an image scaling method that fills the pixels of the interpolated image by selecting the original pixel value closest to the target pixel position. In the continuous domain, we can define the nearest neighbor operator $N(\cdot)$ as follows,
\begin{equation}[N(F)](x,A)=F_{i^{\star}j^{\star}}^{A},\end{equation}    
where $(i^{*},j^{*})=arg\min_{ij}||x_{ij}-x||_{2}^{2}$.

\noindent\textbf{Bilinear Upsampling.}
Bilinear interpolation calculates the new pixel value by taking the weighted average of the four surrounding known pixel values. In the continuous domain, we can define the bilinear interpolation operator $B(\cdot)$ as follows,
\begin{equation}
[B(F)](x,A) = \sum_{i=1}^{2}\sum_{j=1}^{2} \lambda_{ij}f(Q_{ij}),           
\end{equation}
where $\lambda_{ij}$ are the coefficients of bilinear interpolation and $f(Q_{ij})$ represent the grid points, $x=[x_1,x_2]^T\in\mathbb{R}^2$ denotes the 2D spatial coordinates, $x_{1} \in [i,i+1]$, $x_{2} \in [j,j+1]$.

Both of the aforementioned upsampling operators are widely utilized across various network architectures, making it essential to analyze their mathematical properties within rotational equivariant networks. Accordingly, we evaluated their equivariant errors under a rotational equivariant framework.

\begin{Thm}
Assume that a feature map $F \in \mathbb{R}^{n \times n \times t}$ is discretized from the smooth function $e: \mathbb{R}^2 \times S \rightarrow \mathbb{R}$, $|S|=t$, the mesh size is $h$, $U(\cdot)$ is the upsampling operator. 
If for any $A,B \in S, x \in \mathbb{R}^2$, the following conditions are satisfied:
\begin{equation}
\parallel\nabla e(x,A)\parallel\leq G, 
\end{equation}
then the following results are satisfied:
\begin{equation}
\resizebox{\linewidth}{!}{$
|U\left[\tilde{\pi}_{B}^E\right]\left(F\right)\left(x,A\right) - \pi_{B}^E\left[U\left(F\right)\right]\left(x,A\right)| \leq 2(\sqrt{2}+1)Gh.
$}
\end{equation}
\end{Thm}

The upsampling operator $U(\cdot)$ can be either $N(\cdot)$ or $B(\cdot)$. It is worth noting that the above conclusions indicate that the error introduced by the upsampling operator in the rotational equivariant network is related to $h$, aligning with established understanding.

\begin{figure*}[t]
  \centering
  \includegraphics[width=\linewidth]{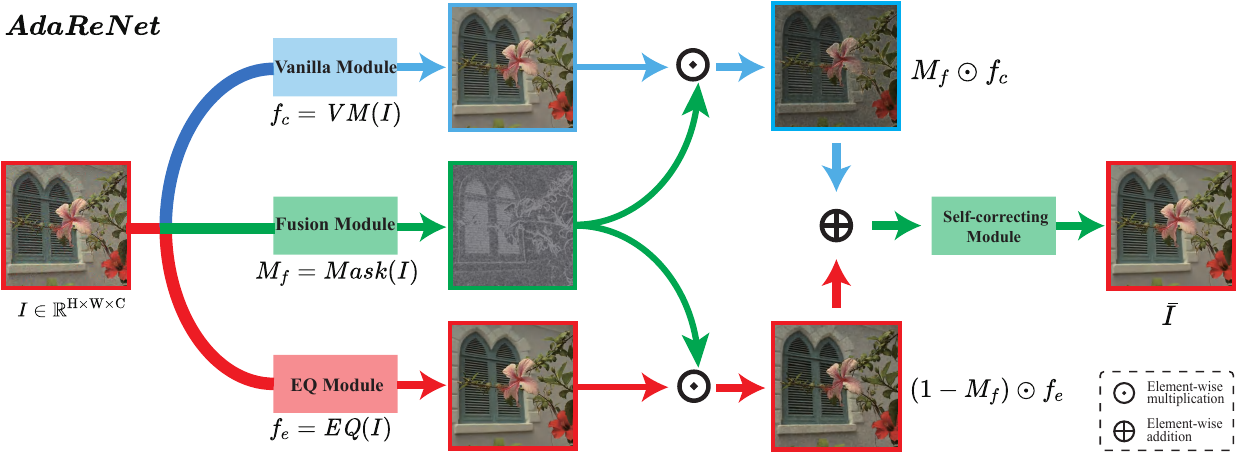}
  \caption{Illustrations of our proposed adaptive network AdaReNet. Specifically, $I \in \mathbb{R}^{\mathrm{H} \times \mathrm{W} \times \mathrm{C}}$ represents a noisy image, where $H$ and $W$ represent the spatial dimensions, and $C$ denotes the channel dimension. The Vanilla Module and EQ Module each produce their respective preliminary denoising results, denoted as $f_c$ and $f_e$. The Fusion Module $Mask(\cdot)$ automatically decides which areas of the image to use more EQ Module would gain more benefit. After adaptive fusion by $Mask(\cdot)$ and correction by the Self-correcting Module $S_c(\cdot)$, the final denoised image $\bar{I}$ is output.}
  \label{fig:backbone}
\end{figure*}

\subsubsection{Analysis of Complete U-Net Network}

We take the network design of the N2N method~\cite{lehtinen2018noise2noise} as an example to give the derivation of the rotation equivariant error for the entire U-Net architecture. As shown in \cref{fig:N2N_UNet}, we decompose the network into multiple upsampling and downsampling blocks. Each downsampling block (DB) consists of one E-Conv layer (Equivariant Convolution) and a downsampling operator, while each upsampling block (UB) is composed of an upsampling operator and two E-Conv layers, we provide the equivariant error for each block and subsequently derive the equivariant error bounds for the complete network.

Theorem 3 demonstrates the equivariance error of the complete U-Net network under discrete angles, and the corollary further provides the equivariance error of the complete U-Net network under any rotation angle. Note that the mesh size of upsampling and downsampling are different, we define the mesh size of the original picture to be $h$, the mesh size after a $\times 2$ downsampling is $2h$, and so on.

\medskip
\begin{Thm}
For an image $X$ with size $H\times W\times n_0,$ and a N-layer rotation equivariant U-Net network $\mathrm{UNet}_{eq}(\cdot)$, whose channel number of the $l^{th}$ layer is $n_l$, rotation equivariant subgroup is $S\leqslant O(2),|S|=t$ , and activation function is set as ReLU. If the latent continuous function of the $c^{th}$ channel of $X$ denoted as $r_c:\mathbb{R}^2\to\mathbb{R},$ and the latent continuous function of any convolution filters in the $l^{th}$ layer denoted as $\phi^l:\mathbb{R}^2\to\mathbb{R}$, $DB_i$ and $UB_i$ represent the downsampling block and the upsampling block, respectively. $\hat{\Psi}$, $\hat{\Phi}$, and $\hat{\Upsilon}$ represent the convolutional layers in the input, middle, and output stages, respectively. We define:
\begin{equation}
\resizebox{\linewidth}{!}{$
    \mathrm{UNet}_{eq}(\cdot)=\hat{\Upsilon}\left[\hat{UB}_{m}\cdots\hat{UB}_{1}\left[\hat{\Phi}\left[\hat{DB}_{m}\cdots\hat{DB}_{1}\left[\hat{\Psi}\right]\right]\cdots\right]\right](\cdot),
$}
\end{equation}
the following conditions are satisfied:
\begin{equation} \label{upper}
\begin{aligned}
&\begin{aligned}|r_{c}(x)|\leq F_{0},\|\nabla_{x}r_{c}(x)\|\leq G_{0},\|\nabla_{x}^{2}r_{c}(x)\|\leq H_{0},\end{aligned} \\
&\begin{aligned}|\phi^l(x)|\leq F_l,\|\nabla_x\phi^l(x)\|\leq G_l,\|\nabla_x^2\phi^l(x)\|\leq H_l,\end{aligned} \\
&\forall\|x\|\geq{(p+1)h/2},\phi_{l}(x)=0,
\end{aligned}
\end{equation}
where $p$ is the filter size, $h$ is the mesh size, $\theta_k = \frac{2k\pi}{t}$, $k = 1,2,\cdots, t$. $\nabla_x$ and $\nabla_x^2$ denote the operators of gradient and Hessian matrix, respectively. We have
\begin{equation}
\left|\mathrm{UNet}_{e q}\left[\tilde{\pi}_{\theta_k}^R\right](X)-\tilde{\pi}_{\theta_k}^R\left[\mathrm{UNet}_{e q}\right](X)\right| \leq R_1h+R_2h^2.
\end{equation}
where $R_1,R_2$ are two constants with respect to $N, n_l$ and the upper bound in \eqref{upper}, their specific values can be found in the supplementary materials.
\end{Thm}

\begin{corollary}
Under the same condition as Theorem 3, for an arbitrary $\theta \in [0, 2\pi]$, let $\pi_{\theta}$ denote the rotation transformation, then $\forall \theta$ we have
\begin{equation}
\resizebox{\linewidth}{!}{$
\left|\mathrm{UNet}_{eq}\left[\tilde{\pi}_{\theta}^R\right](X)-\tilde{\pi}_{\theta}^R\left[\mathrm{UNet}_{e q}\right](X)\right| \leq R_1h+R_2h^2+R_3t^{-1}h,
$}
\end{equation}
where $R_1,R_2,R_3$ are constants that can be found in the supplementary materials.
\end{corollary}

\begin{figure}[h]
  \centering
   \includegraphics[width=\linewidth]{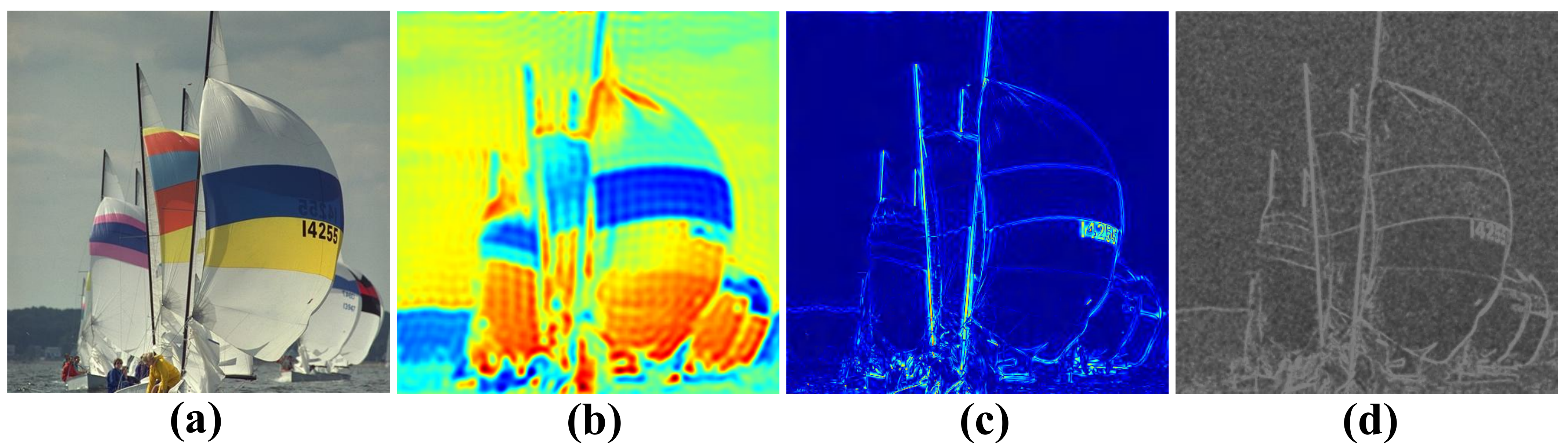}
   \caption{(a) An image from the Kodak dataset, (b) the heatmap of the low-frequency component, (c) the heatmap of the high-frequency component, (d) the output of our proposed MaskNetwork (the brighter area indicates the use of more Vanilla Module).}
   \label{fig:abcd}
\end{figure}

\subsection{Proposed Adaptive Network AdaReNet}Rotation equivariant design usually degrades the representation accuracy of the network because of parameter sharing and filter parameterization and not all areas of natural images strictly comply with rigid rotation equivariance. In order to solve the above problems, the proposed network predominantly comprises four primary modules, as depicted in \cref{fig:backbone}, which are the Vanilla Module, EQ Module, Fusion Module, and Self-correcting Module.

\noindent\textbf{Vanilla Module.} This module utilizes a conventional CNN architecture, maintaining the original design where only translation equivariance is incorporated into the network. 
We define the result of the input image $I$ after passing through the Vanilla Module as $f_c$.
\begin{equation}
    f_c=\mathrm{VM}(I).
\end{equation}
\noindent\textbf{EQ Module.} This module employs the same architecture as the previous one, with the conventional convolution replaced by the rotation equivariant network to incorporate rotation equivariance prior into the architecture. 
We define the result of input image $I$ after passing through the EQ Module as $f_e$, 
\begin{equation}
    f_e=\mathrm{EQ}(I).
\end{equation}

\noindent\textbf{Fusion Module.} This module introduces a MaskNetwork $Mask(\cdot)$, a specialized network designed to merge the outputs from the Vanilla Module and the EQ Module. It employs several layers of standard convolutions to automatically determine which regions of the image would benefit more from a rotation-equivariant network compared to a normal CNN-based network. The output of $Mask(\cdot)$ is shown in \cref{fig:abcd}(d).  
We define the result of input image $I$ after passing through MaskNetwork as $M_f$, 
\begin{equation}
    M_f=\mathrm{Mask}(I).
\end{equation}

\noindent\textbf{Self-correcting Module.} Following the Fusion Module, a self-correcting module is applied to refine the fused result. This is achieved through the use of simple ResNet Blocks. We define the Self-correcting Module as $S_c(\cdot)$.

Based on the network presented in \cref{fig:backbone}, we can derive the following mathematical representation to encapsulate the training and inference processes of the entire network:
\begin{equation}
\hat{I}=M_f\odot f_c + (1-M_f) \odot f_e, \, \Bar{I}=S_c(\hat{I}),
\end{equation}
where $\hat{I}$ denotes the results of adaptive fusion, and $\Bar{I}$ represents the final restoration result after going through the Self-correcting Module, $\odot$ denote element-wise multiplication. 

\noindent\textbf{Loss Function.} We incorporate the loss of the two subnetworks as regularization terms into the main loss. The loss function is defined as follows:
\begin{equation}
\mathrm{L}=\| \overline{I}-\text{target} \|_2+\alpha_1\|f_c-\text{target}\|_2+\alpha_2\|f_e-\text{target}\|_2,
\end{equation}
where $\alpha_1, \alpha_2$ are hyperparameters and we empirically set $\alpha_1=\alpha_2=0.1$.

\noindent\textbf{Remark.}
Embedding rotational equivariance into the network is highly effective for self-supervised image denoising task. Since natural images do not adhere to strict rotational equivariance, our proposed adaptive rotational equivariant network, named AdaReNet, can automatically decide which regions of the image to apply the rotation-equivariant network, thereby further enhancing performance. By observing the output of the Fusion Module in \cref{fig:abcd}(d), it is noted that the mask values are larger at the edge details of the pattern, indicating that the adaptive equivariant network tends to use the outputs from the Vanilla Module more frequently in these areas. Conversely, for the restoration of the majority of low-frequency components in the image, the network predominantly uses the outputs from the EQ Module. This aligns with the common sense that convolutions are more adept at fitting high-frequency information\cite{wang2020high}.
\section{Experiments}
\label{Experiments}

\begin{figure*}[t]
  \centering
  \includegraphics[width=\linewidth]{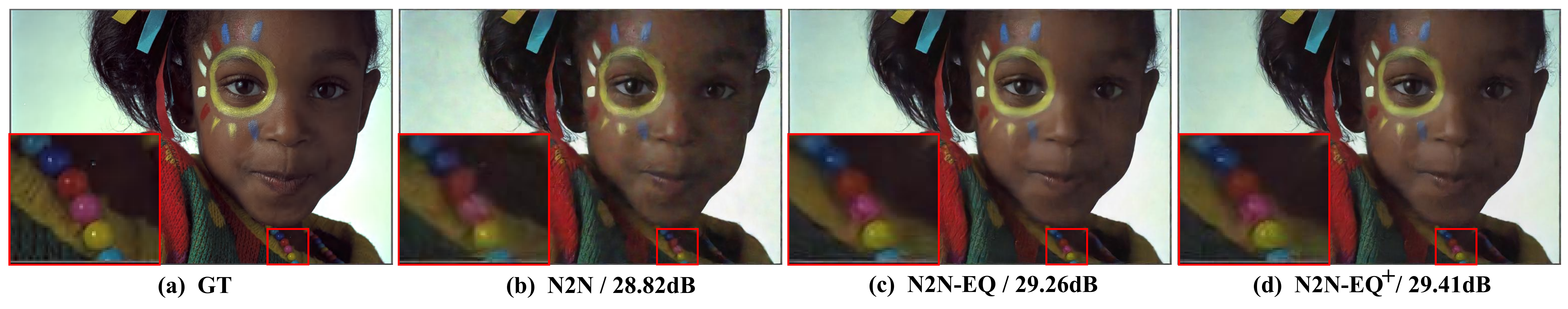}
  \caption{N2N: image denoising results of one image from kodak with $\sigma=50$.}
  \label{fig:N2N_50}
\end{figure*}

In this section, we conducted experiments based on the existing setup and validated our method in three classic approaches. 
For the Noise2Noise~\cite{lehtinen2018noise2noise} and Noise2Void~\cite{krull2019noise2void} methods, 
we showed experiments on U-Net~\cite{ronneberger2015u}, which can significantly speed up the training process while maintaining acceptable performance, and demonstrated the effectiveness of our method.
Besides, we also conducted the R2R experiments based on DnCNN~\cite{pang2021recorrupted}, further validating the superiority of the proposed method. 
Due to space limitations, we only present partial results. Implementation details and further experiments concerning different models, various datasets, model parameter counts, as well as experiments in the field of self-supervised fluorescence microscopy denoising\cite{liu2024fm2s} can be found in supplementary materials.

\subsection{Experiments on Compared Methods}

\textbf{Rotation Equivariant N2N:} We selected the most classical Gaussian noise for our experiments, and randomized the standard deviation $\sigma \in [0, 50]$ of noise for each training example individually. 
N2N-EQ denotes the rotation equivariant network, while $\text{N2N-EQ}^+$ serves as the adaptive rotation equivariant network. The notation in other experiments is similar. The results are shown in \cref{tab:N2N_UNet}.  
The superiority of our method can also be observed from 
\cref{fig:N2N_50}.

\begin{table}[h]
    \centering
    \resizebox{1.0\linewidth}{!}{
    \begin{tabular}{@{}lcccccc@{}}
        \toprule
        \multirow{2}{*}{Dataset} & \multicolumn{3}{c}{Gaussian25} & \multicolumn{3}{c}{Gaussian50} \\
        \cmidrule(r){2-7}
        & N2N~\cite{lehtinen2018noise2noise} & N2N-EQ & $\text{N2N-EQ}^+$ & N2N~\cite{lehtinen2018noise2noise} & N2N-EQ & $\text{N2N-EQ}^+$\\
        \midrule
        Kodak~\cite{franzen1999kodak} & $31.47$/$0.874$ & $31.60$/$0.878$ & $\mathbf{31.72}$/$\mathbf{0.880}$ & $28.29$/$0.778$ & $28.58$/$0.790$ & $\mathbf{28.69}$/$\mathbf{0.791}$ \\
        BSD300~\cite{martin2001database} & $30.18$/$0.869$ & $30.28$/$0.872$ & $\mathbf{30.36}$/$\mathbf{0.873}$ & $27.02$/$0.762$ & $27.24$/$0.771$ & $\mathbf{27.31}$/$\mathbf{0.772}$ \\
        Set14~\cite{zeyde2012single} & $30.02$/$0.851$ & $30.06$/$0.854$ & $\mathbf{30.19}$/$\mathbf{0.855}$ & $27.16$/$0.768$ & $27.32$/$0.775$ & $\mathbf{27.44}$/$\mathbf{0.777}$ \\
        \bottomrule
    \end{tabular}
    }
    \caption{N2N: three networks with U-Net architecture were tested under conditions of Gaussian noise at levels 25 and 50.}
    \label{tab:N2N_UNet}
\end{table}

\begin{figure*}[t]
  \centering
  \includegraphics[width=\linewidth]{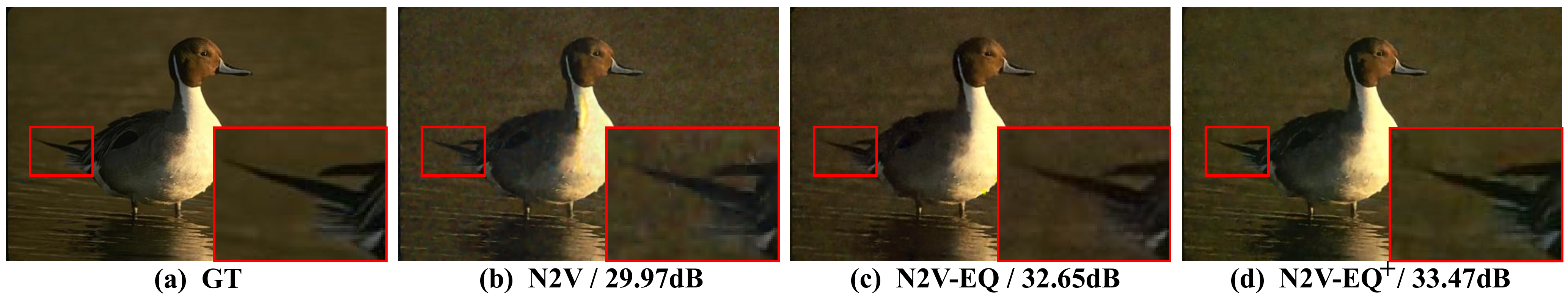}
  \caption{N2V: image denoising results of one image from BSD500 with $\sigma=25$.}
  \label{fig:N2V_25}
\end{figure*}

\vspace{-3mm}
\noindent\textbf{Rotation Equivariant N2V:} We conducted experiments with U-Net architectures. The results are presented in \cref{tab:N2V-UNet} and \cref{fig:N2V_25}. The equivariant error\footnote{$\|[L_R\Phi(f)-\Phi L_R(f)\|_2^2/\|L_R\Phi(f)\|_2^2$, where $\Phi(\cdot)$ represents the network and 
$L_R(\cdot)$ denotes the rotation transformation.} of networks N2V, N2V-EQ, and $\text{N2V-EQ}^+$ are 0.233, 0.068, and 0.076, respectively. This verifies that the improvements are achieved by reducing equivariant errors. Besides, \cref{tab4} further performed diverse scenarios trained on the BSD500 dataset. Our method consistently achieved superior results.

\begin{table}[h]
    \centering
    
    {\Huge
    \resizebox{1.0\linewidth}{!}{
    \begin{tabular}{@{}lcccccc@{}}
        \toprule
        \multirow{2}{*}{Dataset} & \multicolumn{3}{c}{Gaussian25} & \multicolumn{3}{c}{Gaussian50} \\
        \cmidrule(r){2-7}
        & N2V~\cite{krull2019noise2void} & N2V-EQ & $\text{N2V-EQ}^+$ & N2V~\cite{krull2019noise2void} & N2V-EQ & $\text{N2V-EQ}^+$\\
        \midrule
        BSD500~\cite{martin2001database} & 28.17/0.820 & 29.05/0.834 & \textbf{29.12}/\textbf{0.845} & 26.07/0.725 & 26.38/0.735 & \textbf{26.82}/\textbf{0.755} \\
Kodak24~\cite{franzen1999kodak} & 28.86/0.811 & 29.78/0.825 & \textbf{29.93}/\textbf{0.836} & 26.75/0.716 & 27.15/0.732 & \textbf{27.72}/\textbf{0.754} \\
Set14~\cite{zeyde2012single} & 27.22/0.800 & 28.04/0.806 & \textbf{28.09}/\textbf{0.816} & 25.40/0.715 & 25.65/0.726 & \textbf{26.23}/\textbf{0.749} \\
Average & 28.08/0.811 & 28.96/0.822 & \textbf{29.05}/\textbf{0.832} & 26.07/0.719 & 26.39/0.731 & \textbf{26.92}/\textbf{0.753} \\
        \bottomrule
    \end{tabular}
    }}
    \caption{
    N2V: three networks were tested under conditions of Gaussian noise at levels 25 and 50.}
    \label{tab:N2V-UNet}
\end{table}
\vspace{-0.5cm}

\begin{table}[h]
    \centering
    {\huge
    \resizebox{1.0\linewidth}{!}{
    \begin{tabular}{@{}lccccc@{}} 
        \toprule
        Method & Poisson & Poisson30 & Poisson\&Gaussian & Peppersalt & Speckle \\
        \midrule
        N2V & 30.82/0.912 & 32.38/0.961 & 27.91/0.821 & 23.93/0.782 & 26.03/\textbf{0.775} \\
        N2V-EQ(ours) & 31.22/0.904 & 33.59/0.966 & 28.84/0.835 & 24.53/0.792 & \textbf{26.42}/0.736 \\
        $\text{N2V-EQ}^+$(ours) & \textbf{32.19}/\textbf{0.924} & \textbf{35.96}/\textbf{0.976} & \textbf{29.34}/\textbf{0.850} & \textbf{24.83}/\textbf{0.811} & 24.85/0.676 \\
        \bottomrule
    \end{tabular}
    }}
    
    \caption{Quantitative comparison on diverse scenarios.}
    \label{tab4} 
\end{table}

\begin{table*}[t!]
    \scriptsize
    \centering
    
    \begin{tabular}{@{}lcccccccccccccccc@{}}
        \toprule
        \multicolumn{8}{c}{Gaussian25} & \multicolumn{8}{c}{Gaussian50} \\
        \midrule
        \multirow{2}{*}{Method} & \multicolumn{2}{c}{Kodak24~\cite{franzen1999kodak}} & \multicolumn{2}{c}{BSDS300~\cite{martin2001database}} & \multicolumn{2}{c}{Set14~\cite{zeyde2012single}} & \multicolumn{2}{c}{Average} & \multicolumn{2}{c}{Kodak24~\cite{franzen1999kodak}} & \multicolumn{2}{c}{BSDS300~\cite{martin2001database}} & \multicolumn{2}{c}{Set14~\cite{zeyde2012single}} & \multicolumn{2}{c}{Average} \\
        \cmidrule(r){2-17}
          & PSNR & SSIM & PSNR & SSIM & PSNR & SSIM & PSNR & SSIM & PSNR & SSIM & PSNR & SSIM & PSNR & SSIM & PSNR & SSIM \\
        \midrule
        CNN  & 31.47 & 0.874 & 30.18 & 0.869 & 30.02 & 0.851 & 30.56 & 0.865 & 28.29 & 0.778 & 27.02 & 0.762 & 27.16 & 0.768 & 27.49 & 0.769 \\
        G-CNN  & 31.39 & 0.872 & 30.23 & 0.869 & 30.02 & 0.852 & 30.55 & 0.864 & 28.04 & 0.774 & 26.85 & 0.757 & 26.88 & 0.762 & 27.26 & 0.764 \\
        E2-CNN    & 31.23 & 0.869 & 30.02 & 0.864 & 29.70 & 0.845 & 30.32 & 0.859 & 28.04 & 0.768 & 26.86 & 0.754 & 26.83 & 0.758 & 27.24 & 0.760 \\
        PDO-eConv & 31.42 & 0.874 & 30.15 & 0.869 & 29.83 & 0.849 & 30.47 & 0.864 & 28.34 & 0.783 & 27.06 & 0.765 & 27.05 & 0.768 & 27.48 & 0.772 \\
        Fconv & \textbf{31.60} & \textbf{0.878} & \textbf{30.28} & \textbf{0.872} & \textbf{30.06} & \textbf{0.854} & \textbf{30.65} & \textbf{0.868} & \textbf{28.58} & \textbf{0.790} & \textbf{27.24} & \textbf{0.771} & \textbf{27.32} & \textbf{0.775} & \textbf{27.71} & \textbf{0.779} \\
        \bottomrule
    \end{tabular}
    
    \caption{Ablation on rotation-equivariant networks: PSNR and SSIM results for Gaussian25 and Gaussian50.}
    \label{tab:ablation_equivariance}
\end{table*}

\vspace{-3mm}
\noindent\textbf{Rotation Equivariant R2R:} 
The experimental results are shown in \cref{tab:R2R}. The parameter counts for methods R2R, R2R-EQ, and $\text{R2R-EQ}^+$ are 0.67M, 0.17M and 0.84M, respectively. Due to the larger number of adaptive network parameters, it is expected that the performance is inferior to our EQ version when the training dataset is small.

\setlength{\floatsep}{5pt}
\begin{table}[h]
    \centering
    \small  
     \centering \setlength{\tabcolsep}{8.5pt}
    \begin{tabular}{@{}lcccc@{}}  
        \toprule
        Dataset & $\sigma$ & R2R~\cite{pang2021recorrupted} & R2R-EQ & $\text{R2R-EQ}^+$ \\
        \midrule
        BSD68 & 25 & 29.03/0.822 & \textbf{29.14}/0.825 & 29.10/\textbf{0.826}\\
        BSD68 & 50 & 26.01/0.700 & \textbf{26.14/0.711} & 26.00/0.700\\
        \bottomrule
    \end{tabular}
    \caption{
    R2R: three networks were tested under conditions of Gaussian noise at levels 25 and 50.}
    \label{tab:R2R}
\end{table}

\subsection{Ablation Study}

\noindent\textbf{Selection of rotation-equivariant networks:} We conducted ablation experiments with mainstream rotation-equivariant networks, including Cohen \etal~\cite{cohen2016group}, Weiler \etal~\cite{weiler2019general} and Shen \etal~\cite{shen2020pdo}. \cref{tab:ablation_equivariance} confirms that the chosen Fconv~\cite{xie2022fourier} approach is the most effective method in the field of low-level vision.

\begin{table}[t!]
\centering
     {\Huge
    \resizebox{1.0\linewidth}{!}{
\begin{tabular}{@{}lcccccc@{}}
\toprule
\multirow{2}{*}{Dataset} & \multicolumn{3}{c}{Gaussian25} & \multicolumn{3}{c}{Gaussian50} \\
\cmidrule(r){2-7}
& N2V~\cite{krull2019noise2void} & N2V-EQ & $\text{N2V-EQ}^+$ & N2V~\cite{krull2019noise2void} & N2V-EQ & $\text{N2V-EQ}^+$\\
\midrule
BSD500~\cite{martin2001database} & 23.09/0.758 & 28.19/0.811 & \textbf{29.05}/\textbf{0.847} & 22.20/0.663 & 26.06/0.718 & \textbf{26.47}/\textbf{0.745} \\
Kodak24~\cite{franzen1999kodak} & 22.77/0.758 & 28.98/0.809 & \textbf{29.82}/\textbf{0.840} & 22.72/0.657 & 26.96/0.721 & \textbf{27.32}/\textbf{0.743} \\
Set14~\cite{zeyde2012single} & 22.03/0.736 & 27.07/0.787 & \textbf{28.04}/\textbf{0.820} & 19.65/0.609 & 25.27/0.707 & \textbf{25.67}/\textbf{0.731} \\
Average & 22.63/0.751 & 28.08/0.802 & \textbf{28.97}/\textbf{0.836} & 21.52/0.643 & 26.10/0.715 & \textbf{26.49}/\textbf{0.740} \\
\bottomrule
\end{tabular}
}}

\caption{N2V w/o rotation augmentation: three networks were tested under conditions of Gaussian noise at levels 25 and 50.}
\label{tab:N2V_rot_ablation}
\end{table}

\noindent\textbf{Data rotation augmentation:} Rotational data augmentation on training data is a commonly used method to enhance model performance and robustness. We conducted ablation experiments with rotation augmentation in the N2V method. \cref{tab:N2V-UNet} shows the experiments on the U-Net network with rotation augmentation, while \cref{tab:N2V_rot_ablation} presents our ablation experiments without augmentation.

\section{Conclusion}
\label{Conclusion}
\vspace{-2mm}
In this work, we first explore and introduce rotation equivariant image prior into the self-supervised image denoising task at the network architecture level. Through rigorous theoretical analysis, we prove that simply replacing the convolution layers with ECNNs leads to a rotational equivariant version of the network. Building on this, we propose AdaReNet, an adaptive rotation equivariant network, which further enhances performance. 
Extensive experiments demonstrate the effectiveness of our approach, confirming that incorporating rotation equivariant prior significantly improves denoising results. Overall, our work underscores the importance of leveraging rotation invariance for self-supervised learning and sets a foundation for future research in this field.

\noindent\textbf{Acknowledgement.} This research was supported by NSFC project under contract U21A6005; the Major Key Project of PCL under Grant PCL2024A06; Tianyuan Fund for Mathematics of the National Natural Science Foundation of China (Grant No.12426105) and Key Research and Development Program (Grant No. 2024YFA1012000).

{
    \small
    \bibliographystyle{ieeenat_fullname}
    \bibliography{main}
}


\end{document}